\documentclass{article}

\usepackage{microtype}
\usepackage{graphicx}
\usepackage{subfigure}
\usepackage{booktabs}
\usepackage{hyperref}

\usepackage[accepted]{icml2025}

\usepackage{amsmath}
\usepackage{amssymb}
\usepackage{mathtools}
\usepackage{amsthm}

\usepackage[capitalize,noabbrev]{cleveref}

\theoremstyle{plain}

\theoremstyle{definition}

\theoremstyle{remark}

\usepackage[textsize=tiny]{todonotes}

\icmltitlerunning{Label Space Reduction for Zero-shot Classification}

\begin{document}

\twocolumn[
\icmltitle{From Haystack to Needle: Label Space Reduction for Zero-shot Classification}

\icmlsetsymbol{equal}{*}

\begin{icmlauthorlist}
\icmlauthor{Nathan Vandemoortele}{equal,comp}
\icmlauthor{Bram Steenwinckel}{comp}
\icmlauthor{Femke Ongenae}{comp}
\icmlauthor{Sofie {Van Hoecke}}{comp}
\end{icmlauthorlist}

\icmlaffiliation{comp}{IDLab, Ghent University - imec, Gent, Belgium}

\icmlcorrespondingauthor{Nathan Vandemoortele}{nathan.vandemoortele@ugent.be}

\icmlkeywords{Large Language Models, Zero-shot, Classification, Label Space}

\vskip 0.3in
]

\printAffiliationsAndNotice{}

\begin{abstract}
We present Label Space Reduction (LSR), a novel method for improving zero-shot classification performance of Large Language Models (LLMs). LSR iteratively refines the classification label space by systematically ranking and reducing candidate classes, enabling the model to concentrate on the most relevant options. By leveraging unlabeled data with the statistical learning capabilities of data-driven models, LSR dynamically optimizes the label space representation at test time. Our experiments across seven benchmarks demonstrate that LSR improves macro-F1 scores by an average of 7.0\% (up to 14.2\%) with \textsc{Llama-3.1-70B} and 3.3\% (up to 11.1\%) with \textsc{Claude-3.5-Sonnet} compared to standard zero-shot classification baselines. To reduce the computational overhead of LSR, which requires an additional LLM call at each iteration, we propose distilling the model into a probabilistic classifier, allowing for efficient inference.
\end{abstract}

\section{Introduction}
Zero-shot learning is a powerful paradigm that enables machine learning models to handle new tasks without requiring task-specific training. Instead, these models transfer knowledge from previously learned tasks to tackle novel challenges~\cite{wang2019survey}. A key application of this paradigm is zero-shot classification. In this approach, models can identify examples from classes they've never seen during training by making connections to known classes through semantic embeddings or attribute spaces~\cite{Xian2017ZeroShotLC,wang2019survey}. This capability addresses a major limitation of traditional supervised learning: the need for labeled training data for every target class. As a result, zero-shot classification is particularly valuable in real-world applications where collecting labeled examples for all possible classes is expensive or impractical~\cite{Xian2017ZeroShotLC}.

The field of Large Language Models (LLMs) has further advanced zero-shot classification by leveraging their extensive pretraining on diverse datasets. Consequently, LLMs demonstrate an unprecedented ability to understand and generate human-like text, enabling them to infer semantic relationships between input text and class descriptions without task-specific fine-tuning \cite{brown2020language,Bommasani2021FoundationModels,wei2022finetuned}.
For example, when presented with a maintenance log entry like ``Bearing temperature increased to 85°C with unusual vibration patterns and metallic noise during operation," an LLM can identify critical indicators such as ``increased temperature," ``vibration," and ``metallic noise" to accurately classify the text under the ``bearing failure" category, even if equipment failure classification was not explicitly included in its pretraining. Such versatility highlights the practical utility of LLMs in real-world applications.

However, current approaches to LLM zero-shot classification often rely on simplistic methods that present class options as unstructured, flat lists within the prompt. This becomes increasingly problematic as the number of potential classes expands. Specifically, LLMs face critical limitations when dealing with long contexts due to key limitations, often attributed to factors such as positional bias~\cite{Peysakhovich2023AttentionSC,An2024WhyDT} and attention dilution~\cite{Peysakhovich2023AttentionSC}, reducing their ability to accurately distinguish between a large set of potential classes. As context length grows, the fixed attention budget of transformers must be distributed across more tokens, as the attention weights are normalized to sum to 1. This shared attention makes it harder to focus strongly on relevant information, as demonstrated in ``Needle in the haystack" tests~\cite{kamradt2023needle,hsieh2024ruler} where LLMs struggle to retrieve key facts from long contexts~\cite{hsieh2024ruler,liu2023lostmiddlelanguagemodels}. The challenge becomes even more complex when multiple ``needles" must not only be retrieved, but also reasoned over~\cite{langchain2023multineedle,hsieh2024ruler}. As a result, by reducing the label space, LLMs can allocate their attention more effectively, enhancing reasoning processes such as step-by-step thinking ~\cite{wei2023chainofthoughtpromptingelicitsreasoning,kojima2023largelanguagemodelszeroshot}, where each step builds upon previous information~\cite{anonymous2025retrieval}.

Reducing the label space is not a new concept and is well-established in multi-label classification, generally falling into two categories: those that leverage correlations between features and labels through machine learning models~\cite{group2019,labelspacedimension2020}, and those that operate independently of the feature space~\cite{moyano2022}. Feature-independent approaches include methods such as label embeddings~\cite{tai2012} or are based on hierarchical or descriptive label relationships~\cite{Charte2014LIMLCAL}. However, all these approaches have limitations as they typically require labeled training data and are not model-agnostic~\cite{moyano2022}, and thus cannot be readily applied. Moreover, they generate unordered static label spaces that cannot dynamically adapt to generative outputs of LLMs. While the concept of label space reduction has been implicitly demonstrated in RAG systems~\cite{gao2024retrievalaugmentedgenerationlargelanguage}, addressing similar challenges, label space reduction strategies for LLMs in the context of zero-shot classification remain, to our knowledge, an underexplored area.

In this paper, we introduce Label Space Reduction (LSR), a novel approach to tackle the challenges of large label spaces in zero-shot multi-classification tasks. The presented method makes two key contributions: First, we develop an innovative iterative system that refines the classification label space by ranking and reducing the label space, significantly improving zero-shot multi-classification accuracy by enabling LLMs to focus on the most relevant options. Second, we propose a cost-efficient solution that distils the LSR process into a probabilistic classifier, addressing the computational overhead of LSR while preserving classification accuracy and enabling efficient inference.

\section{Related Work}\label{sec:related_work}
Our work builds upon three key research areas, namely zero-shot classification with LLMs, extreme multi-label classification, and sampling strategies used in language model decoding.

\subsection{Zero-shot Classification with LLMs}
While initially developed for text classification, LLMs have demonstrated remarkable zero-shot classification capabilities across diverse domains, including image classification through vision-language models~\cite{radford2021learning,alayrac2022flamingo}, tabular data classification~\cite{hegselmann2023tabllm}, time series classification~\cite{zhang_large_2024}, and even audio classification~\cite{latif2023sparkslargeaudiomodels}. Early work demonstrated that providing task descriptions and class definitions in natural language could guide models to make reasonable predictions~\cite{kojima2022large,wei2022chain}. This capability was further enhanced through prompt engineering techniques~\cite{Sahoo2024ASS} and the development of instruction-tuned models~\cite{chung2022scaling}. While these approaches have shown promising results, their performance can be sensitive to prompt design and may struggle with nuanced category distinctions. More recent work has focused on improving model reasoning through popular techniques like chain-of-thought reasoning~\cite{wei2023chainofthoughtpromptingelicitsreasoning,kojima2023largelanguagemodelszeroshot} and combining robustness methods such as self-consistency~\cite{wang2023selfconsistencyimproveschainthought}. However, these approaches do not fundamentally address the challenges posed by large-scale classification tasks with long context requirements.

\subsection{Extreme Multi-label Classification}
There are notable connections between our work and extreme multi-label classification (XMC), a field that addresses classification problems involving thousands to millions of classes~\cite{bhatia2015,mittal_multi-modal_2022,zhu2024icxmlincontextlearningframework}. While XMC tasks have traditionally been approached in a supervised manner, researchers have also explored zero-shot XMC methods~\cite{chang2020taming}. Early approaches relied on retrieval-based techniques such as TF-IDF~\cite{salton1988term} and BM25~\cite{robertson2009probabilistic} to match input texts with label embeddings. A major advancement occurred when \citet{reimers2019sentence} demonstrated the effectiveness of dense embeddings for retrieval, which were subsequently applied to various XMC tasks~\cite{you2019attentionxml,chang2020taming,jiang2021lightxml}. More recent work has focused on LLM-based approaches, emphasizing true zero-shot settings where test labels are entirely unseen, unlike generalized zero-shot settings that may include seen labels. For example, ICXML~\cite{zhu2024icxmlincontextlearningframework} adopts a hybrid generation-and-retrieval approach, first generating demonstrations with an LLM and then aligning the outputs with available labels to create a shortlist of candidates. LMTX~\cite{Zhang2024ZeroShotLO} achieved state-of-the-art performance by employing an LLM as a teacher to iteratively train a bi-encoder. However, these approaches still incorporate retrieval-based components and impose an arbitrary label-space cut-off, effectively constraining the LLM’s performance to that of the underlying embedding model. Moreover, we propose a solution that prioritize the role of the LLM as a reasoner for classification, which is applicable to solving classification tasks beyond just text classification. 

\subsection{Sampling Strategies}
Language model decoding is the problem of generating text by selecting tokens from the model’s predicted probability distribution. Early methods like greedy selection and random sampling often led to repetitive or incoherent outputs. Beam search and temperature-based sampling~\cite{temperature2017} improved diversity and quality but failed to address the long tail of irrelevant and low-probability tokens, which can collectively hold significant probability mass~\cite{Holtzman2020The}.
This realization led to the development of more distribution-aware sampling strategies. Top-k sampling~\cite{fan2018hierarchicalneuralstorygeneration} limits choices to the k most probable tokens, while nucleus (Top-p) sampling~\cite{Holtzman2020The} selects tokens whose cumulative probability exceeds p. More recently, Min-p sampling~\cite{nguyen2024turningheatminpsampling} enforces a minimum probability threshold to maintain coherence. Our implementation leverages these distribution-aware thresholds to filter out low-probability classes.

\section{Methodology}\label{sec:methodology}
\begin{figure}[ht]
\vskip 0.1in
\begin{center}
\centerline{\includegraphics[width=\columnwidth]{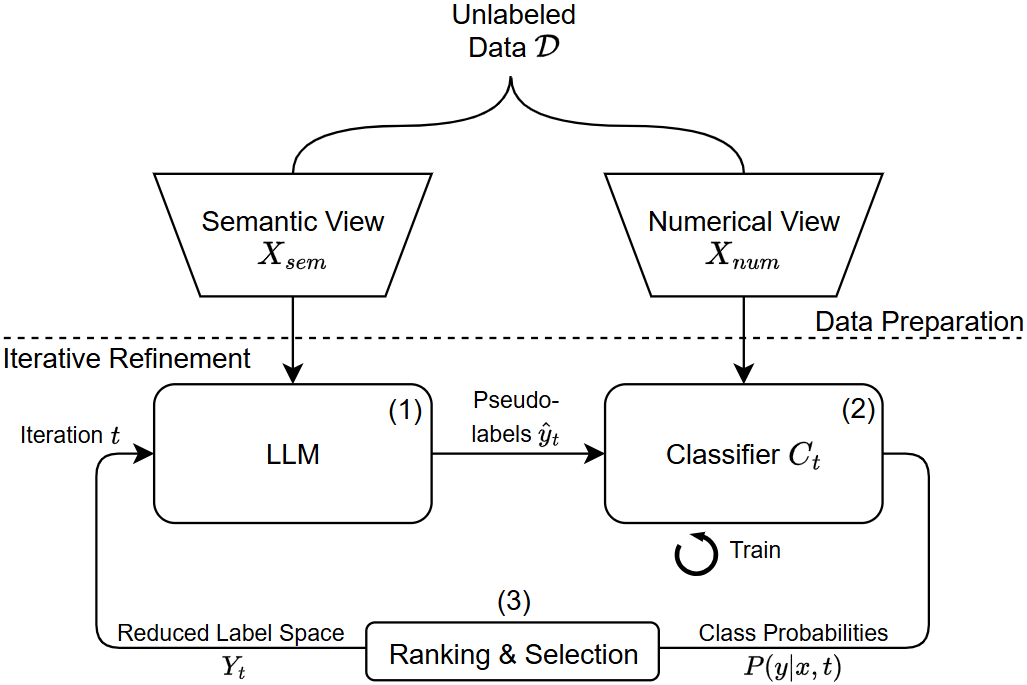}}
\caption{Illustration of the proposed methodology. (1)
The LLM categorizes the data by selecting labels from the full label set. (2) With these pseudo-labels a classifier is trained on the data's numerical representation, which generates probabilities for each class. (3) The labels are then ranked and filtered based on an adaptive threshold to form a reduced label set, which is fed back to the LLM, initiating the next iteration.}
\label{fig:architecture}
\end{center}
\vskip -0.2in
\end{figure}

Addressing shortcomings of current methods, we present our LSR methodology, which consists of three distinct phases that operate in an iterative process. In this section, we detail each phase of the process, as illustrated in Figure~\ref{fig:architecture}.

\subsection{Problem Definition}
Let $\mathcal{X}$ denote the input space and $\mathcal{Y} = \{1,\ldots,K\}$ be a finite label space comprising $K$ distinct classes. Given an unlabeled dataset $\mathcal{D} = \{x_1,\ldots,x_n\}$ where $x_i \in \mathcal{X}$ consisting of $n$ samples, we aim to improve the zero-shot classification performance of an LLM for these $n$ samples, considering the following constraints:
\begin{itemize}
    \item \textbf{Single-label Multi-Classification} Each sample $x_i \in \mathcal{D}$ corresponds to exactly one class-label $y_i \in \mathcal{Y}$
    \item \textbf{Closed-world Assumption} The complete set of possible unseen classes $\mathcal{Y}$ is known and fixed a priori
    \item \textbf{Unlabeled Data} Access to a collection of unlabeled samples $\mathcal{D}$ which can be used for training
\end{itemize}

These constraints represent a common and practical set of conditions found in many real-world zero-shot classification scenarios. The key challenge lies in optimizing classification performance without labeled examples or external knowledge (such as label descriptions or hierarchical relationships), while minimizing token usage and enabling deployment on consumer hardware.

\subsection{Data Representation and Preprocessing}
The framework operates on dual representations of the data:
\begin{enumerate}
\item Numerical view: $X_{num} \in \mathbb{R}^{n\times d}$, where $d$ is the feature dimensionality
\item Semantic view: $X_{sem} = \{s_1,...,s_n\}$, where $s_i$ is the natural language description of $x_i$
\end{enumerate}

The aim of this dual representation approach is to find a statistical correlation between an LLM's predictions and the underlying feature distributions of a novel task. This framework can be applied to classification tasks with arbitrary data types, provided that the data can be represented both numerically and described semantically in natural language.
\subsection{Initial Classification}
The LLM starts generating initial pseudo-labels $\hat{y}_0$ through:
\begin{equation}
\hat{y}_{0} = \text{LLM}(x_{sem}, \mathcal{Y}, prompt_{template})
\end{equation}
where $prompt_{template}$ includes Chain-of-Thought (CoT) reasoning~\cite{wei2023chainofthoughtpromptingelicitsreasoning,kojima2023largelanguagemodelszeroshot} which improves preformance and makes better use of the reduced label space. Initially, all candidate labels are presented as a list of options in the prompt, and the model is tasked to categorize the sample. The full prompt template can be found in Appendix \ref{app:prompt_template}.

\subsection{Probabilistic Classifier}
Based on these pseudo-labels, we train a classifier $C$ on $(X_{num}, \hat{y}_0)$ that outputs probability distributions over classes:
\begin{equation}
\begin{gathered}
P(y|x) = C(x_{num}), \\ 
\text{ where } P(y|x) \in [0,1]^K \text{ and } \sum_i P(y_i|x) = 1
\end{gathered}
\end{equation}
The model is first trained using a hold-out validation strategy to prevent overfitting, then retrained on the complete dataset using the same hyperparameters.\\
By training on the LLM's pseudo-labels, we measure the agreement between predictions based on the underlying feature distributions, effectively quantifying the model's certainty for each class.

\subsection{Ranking and Selection}\label{eq:selection}
For any given sample $x \in D$, the labels are ranked based on their predicted probabilities according to $R(x)$:
\begin{equation}\label{eq:minp}
R(x) = \text{argsort}_{descending}(P(y|x))
\end{equation}

Next, the labels are selected based on the adaptive threshold following Min-p sampling, which includes all labels whose probability exceeds a fraction $p$ of the maximum probability:
\begin{equation}
\text{Min-p}(x, p) = {y \in Y : P(y|x) \geq p \cdot \max(P(y|x))}
\end{equation}
We propose an adapted version of Min-p, denoted as Min-p+, which includes the LLM's current prediction $\hat{y}_t$ at iteration $t$ during the refinement process:
\begin{equation}
\text{Min-p+}(x, p, t) = \text{Min-p}(x, p) \cup {\hat{y}_t}
\end{equation}
The threshold $p$ is optimized to achieve a target average dimensionality $k$ in the reduced label space. Since $p$ is an indirect parameter that affects the final dimensionality, it is more intuitive to control $k$ directly. This relationship can be expressed as:
\begin{equation}
\frac{1}{n} \sum_{i=1}^n w_i|\text{Min-p+}(x_i, p, t)| \approx k
\end{equation}
where $w_i$ is a class-balancing weight inversely proportional to the frequency of $\hat{y}_t$. While $k$ represents an average, the number of candidate labels adapts dynamically per sample, where instances with higher uncertainty naturally receive more candidate labels and vice versa. A detailed comparison of thresholding on the label space following different sampling strategies is provided in Appendix~\ref{app:sampling}.

\subsection{Iterative Refinement}
Our iterative refinement approach, outlined in Algorithm \ref{alg:iterative_refinement}, consists of multiple steps that progressively improve classification accuracy. The process begins by obtaining initial predictions from the LLM (line 4), which are used as pseudo-labels to train a baseline classifier (line 8). In each subsequent iteration, we create a reduced label space by integrating the classifier's confidence scores with the LLM's previous predictions (lines 9-11). The LLM then generates new predictions (line 6), but instead of considering all possible labels, it focuses on a filtered set of ranked candidates for each sample. This filtered approach enables more focused CoT reasoning by limiting the LLM's attention to fewer plausible options. The classifier is then retrained using these refined predictions (line 8). The process continues until reaching a predetermined 
number of iterations $i$ (lines 2-13), where a higher number of iterations typically leads to convergence, at which point no significant changes can be observed in the label space. This iterative process creates a feedback loop where the LLM and classifier help improve each other's performance and potentially correct initial misclassifications with more focused reasoning.

The final predictions $\hat{y}_{\text{final}}$ are determined through majority voting (line 14):
\begin{equation}
    \hat{y}_{\text{final}} = \text{mode}(\{\hat{y}_0,...,\hat{y}_i\})
\end{equation}
By aggregating predictions from previous iterations, we provide robustness against individual iteration errors. Furthermore, aggregation helps mitigate issues when the process fails to reach proper convergence and circumvents the need for complex stopping criteria.

\begin{algorithm}[tb]
\caption{Iterative Label Space Refinement}
\label{alg:iterative_refinement}
\begin{algorithmic}[1]
\REQUIRE Dataset $D$, Label space $Y$, Dimensionality $k$, Iterations $i$
\ENSURE Final predictions $\hat{y}_{\text{final}}$
\STATE predictions $\leftarrow$ [ ]
\FOR{$t = 0$ {\bfseries to} $i$}
    \IF{$t = 0$}
        \STATE $\hat{y}_t \leftarrow \text{LLM}(D, Y)$
    \ELSE
        \STATE $\hat{y}_t \leftarrow \text{LLM}(D, Y_{t-1})$
    \ENDIF
    \STATE $C_t \leftarrow \text{TrainClassifier}(D, \hat{y}_t)$
    \STATE $P_t \leftarrow C_t(D)$
    \STATE $p \leftarrow \text{FindOptimalThreshold}(P_t, k, \hat{y}_t)$
    \STATE $Y_t \leftarrow \text{FilterLabelSpace}(P_t, p, \hat{y}_t)$
    \STATE predictions.append$(\hat{y}_t)$
\ENDFOR
\STATE $\hat{y}_{\text{final}} \leftarrow \text{mode}(\text{predictions})$
\end{algorithmic}
\end{algorithm}

\subsection{Practical Inference} \label{sec:practical_inference}
While the complete iterative methodology described above is valuable for offline evaluation and benchmarking, repeated training cycles during live inference may be impractical for real-world applications. For example, in a system monitoring customer support tickets that need to be classified and routed every few minutes, frequent model retraining would introduce unacceptable costs and latency. Therefore, we propose two alternative strategies:

\textbf{Full Inference} This approach reuses the saved classifiers from every iteration to generate rankings, following the established methodology. While this maintains the benefits of LSR and reduces computational overhead, it has two main limitations: the classifier probabilities are not calibrated with new test samples, and each iteration requires an LLM call. Due to these constraints, we do not explore this approach further in this paper.

\textbf{Direct Inference} Alternatively, we propose to train a final classifier $C_{final}$ on the combined LLM predictions from all iterations, effectively distilling an LLM ensemble into a single classifier that remains frozen during deployment.
\begin{equation}
C_{final} = train(X_{num},[{\hat{y}_0,...,\hat{y}_i}])
\end{equation}
For a new sample $x$, we can then directly use the classifier's probabilities to make predictions:
\begin{equation}
\begin{gathered}
P(y|x) = C_{final}(x)\\
\hat{y}_{\text{final}} = \underset{y}{\operatorname{argmax}} (P(y|x))
\end{gathered}
\end{equation}
Our experiments show that making an additional LLM prediction on the reduced set of labels from $C_{final}$ (using Min-p sampling from Equation~\ref{eq:minp} and class-weighting based on the classifier's predictions) only improves performance when dealing with undersampled classes. Thus, direct inference only requires a trained classifier for inference, which allows for the use of larger LLMs or additional test-time scaling techniques during the iterative training phase, enhancing performance at a one-time fixed cost.

\section{Experimental Setup}\label{sec:experimental_setup}
In this section we describe the datasets, evaluation metrics and implementation details used in our experiments.

\subsection{Datasets}
We evaluate our method on seven diverse benchmark datasets, including five text classification datasets from the MTEB benchmark~\cite{muennighoff2023mtebmassivetextembedding} (\textsc{Amazon Massive Scenario}, \textsc{Amazon Massive Intent}, \textsc{Banking77}, \textsc{Mtop Domain}, \textsc{Mtop Intent}), a text classification dataset (\textsc{DBpedia\footnote{\url{https://www.kaggle.com/datasets/danofer/dbpedia-classes}}}), and a tabular classification dataset (\textsc{Crime\footnote{Police Department Incident Reports: Historical 2003 to May 2018, \url{https://datasf.org/opendata/}}}). We refer to Appendix \ref{app:dataset_characteristics} for detailed descriptions. All datasets contain a substantial number of classes (ranging from 11 to 102) with medium-sized sample sets (2974-4386 samples).

\subsection{Model Selection and Configuration}
\textbf{Large Language Models}
Our framework primarily uses \textsc{Llama-3.1-70B-Instruct}~\cite{grattafiori2024llama3herdmodels}, accessed through the DeepInfra API\footnote{\url{https://deepinfra.com}}, due to its reproducibility and open-weight nature, which enables verification and replication of our results. We selected this model and other recent open-weight models including \textsc{Gemma-2-27b-it}~\cite{gemmateam2024gemma2improvingopen} and \textsc{Qwen-2.5-72B-Instruct}\cite{qwen2.5}, as well as the closed-source \textsc{Claude-3.5-Sonnet(old)}~\cite{anthropic2024claude}, to ensure a diverse representation of model architectures and training approaches. Classification is performed in batches of 10 randomly selected samples, and we instruct the LLM to apply CoT reasoning. To ensure deterministic behavior, we set the temperature parameter to 0. The template for the prompt is shown in Appendix \ref{app:prompt_template}.

\textbf{Text Embeddings} To obtain numerical values for the classification tasks, dense text embeddings are generated using the open-source \textsc{BGE-large-en-v1.5}~\cite{bge2023} Sentence-BERT (SBERT) model (355M parameters, 1.34GB serialized). Through binary quantization in Sentence Transformers~\cite{reimers-2019-sentence-bert}, we compress the vectors from 1024 to 128 dimensions to speed up classifier training.

% 445KB serialized for amazon_massive_intent dataset
\textbf{Classifier} For classification, we utilize \textsc{CatBoost}~\cite{NEURIPS2018_14491b75}, a gradient boosting framework that demonstrates state-of-the-art performance on structured data~\cite{Borisov2021DeepNN}, while being light-weight ($<$1MB when serialized). We use default parameters with $lr=0.1$ and $depth=3$. To mitigate overfitting, we use a stratified hold-out validation set, reserving 20\% of the data for early stopping while maintaining the original class distribution. Class imbalance is addressed through class weights during training. After determining the optimal number of iterations through early stopping criteria, we retrain the classifier on the complete dataset.

Both \textsc{BGE-large-en-v1.5} and \textsc{Catboost} models were evaluated on standard consumer-grade computing hardware (Intel Core i7-1365U processor @ 5.2 GHz, 16GB DDR4 RAM, running Windows 11).

\subsection{Data Preparation}
Our methodology requires both numerical and semantic representations of the data. Here we detail the preparation process for each view.

\textbf{Numerical View} For text classification tasks, numerical text embeddings are generated using \textsc{BAAI/bge-large-en-v1.5}, which produces 128-dimensional feature vectors as input to the classifier. For tabular classification tasks using the \textsc{Crime} dataset, which contains both semantic and numerical features, we treat all semantic features as numerical through categorical encoding. Additionally, we extract the hour of the day from the Time variable as a new feature. In total we obtain 6 features (Description, Police District, Resolution, Address, Day of Week, Hour of Day).

\textbf{Semantic View} To create the semantic representation, we convert all contextual features, whether originally semantic or numerical, into a list of key-value pairs that can be directly incorporated into the prompt. Examples of these representations are provided with the prompt template in Appendix \ref{app:prompt_template}.

\section{Metrics and Evaluation}
Our goal is to measure classification performance, where we assume all classes are of equal importance. Therefore, we evaluate our method using the Macro F1-score, as it treats all classes equally, regardless of their frequency in the dataset:
\begin{equation}
F1_{macro} = \frac{1}{K} \sum_{i=1}^K \frac{2(\text{precision}_i \cdot \text{recall}_i)}{\text{precision}_i + \text{recall}_i}
\end{equation}
where $K$ is the number of classes, and $precision_i$ and $recall_i$ are computed for each class $i$.

It's important to note that while our method involves training during test time, this training occurs exclusively on pseudo-labels generated by the LLM, maintaining the zero-shot nature of our approach. The reported scores are  computed against the ground truth labels of the test set.

\section{Experiments}\label{sec:experiments}
In this section, we present experimental results and key findings of our proposed method. Each experiment was conducted only once to limit expenses without compromising insights.
\subsection{Label Space Reduction}
We analyze how varying the label space size (k) affects the performance of \textsc{Llama-3.1-70B}
across multiple datasets. We measure performance using Macro F1-scores, calculated through majority voting over 15 iterations. For consistency, each experiment with different $k$ values begins with the same zero-shot LLM predictions, starting from randomly shuffled labels in the prompt. Our results follow the complete experimental methodology outlined in Section~\ref{sec:methodology}.

\begin{figure}[ht]
\vskip 0.1in
\begin{center}
\centerline{\includegraphics[width=\columnwidth]{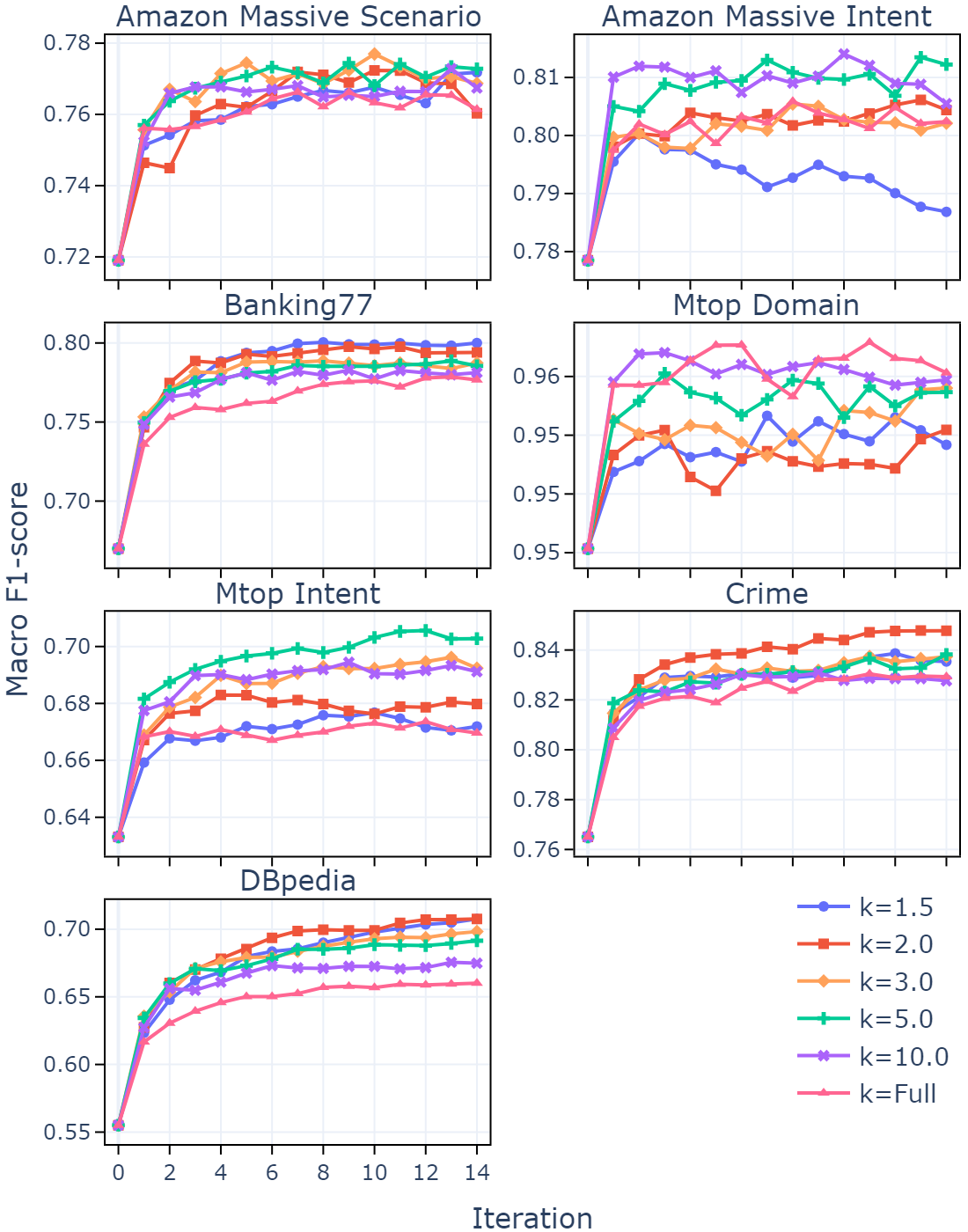}}
\caption{Performance comparison of LSR with varying label space sizes (k). When $k=Full$, the labels are ranked but not reduced. Lines show macro-F1 scores over 15 iterations, starting from zero-shot baseline predictions (\textsc{Llama-3.1-70B}).}
\label{fig:search_space}
\end{center}
\vskip -0.3in
\end{figure}

\begin{table*}[t]
\caption{Results of LSR with varying label
space dimensionality (k). All macro-F1 scores show gains over zero-shot baseline predictions (\textsc{Llama-3.1-70B}). Chain-of-Thought with Self-Consistency (CoT-SC) scores are a majority vote of 15 resamples.}
\label{tab:lsr}
\begin{center}
\begin{small}
\begin{sc}
\begin{tabular}{l|ccccccc|c}
\toprule
& \begin{tabular}[t]{@{}c@{}} \textbf{Amazon} \\ \textbf{Massive} \\ \textbf{Scenario} \end{tabular} & \begin{tabular}[t]{@{}c@{}} \textbf{Amazon} \\ \textbf{Massive} \\ \textbf{Intent} \end{tabular} & \textbf{Banking77} & \begin{tabular}[t]{@{}c@{}} \textbf{Mtop} \\ \textbf{Domain} \end{tabular} & \begin{tabular}[t]{@{}c@{}} \textbf{Mtop} \\ \textbf{Intent} \end{tabular} & \textbf{Crime} & \textbf{DBpedia} & \begin{tabular}[t]{@{}c@{}} 
\\ Average \\ Gain \end{tabular}\\
\midrule
Llama-3.1-70B & .719 & .778 & .670 & .945 & .633 & .765 & .555 \\
\quad w/ CoT-SC& +.015 & +.020 & +.026 & +.011 & +.017 & +.002 & +.024 & +0.016\\
\quad w/ LSR & & & & & & & & \\
\quad\quad k=1.5 & +.052 & +.018 & \textbf{+.129} & +.014 & +.043 & +.067 & +.132 & +0.065 \\
\quad\quad k=2.0 & +.050 & +.027 & \textbf{+.129} & +.014 & +.050 & \textbf{+.077} & \textbf{+.142}  & \textbf{+0.070}\\
\quad\quad k=3.0 & \textbf{+.060} & +.024 & +.121 & +.015 & +.059 & +.069 & +.131  & +0.068\\
\quad\quad k=5.0 & +.058 & +.034 & +.118 & +.016 & \textbf{+.067} & +.068 & +.131  &  \textbf{+0.070}\\
\quad\quad k=10.0 & +.049 & \textbf{+.035} & +.115 & +.016 & +.059 & +.067 & +.117  & +0.065 \\
\quad\quad k=Full & +.044 & +.028 & +.101 & \textbf{+.017} & +.037 & +.061 & +.099 & +0.055\\
\bottomrule
\end{tabular}
\end{sc}
\end{small}
\end{center}
\end{table*}

\textbf{Iterations} Our analysis reveals two key observations from Figure \ref{fig:search_space}. First, the macro-F1 scores show an initial boost after the first iteration, with the most significant gains occurring within the first five iterations, followed by convergence around the tenth iteration. This early convergence pattern suggests that, while our framework supports extended iterations, substantial performance benefits can be achieved with minimal computational overhead. This behavior is consistent across all values of $k$, with the \textsc{Amazon Massive Intent} dataset ($k=1.5$) being a notable exception.\\
Second, we observe distinct performance patterns between different types of datasets. Datasets with rare classes, such as \textsc{Amazon Massive Scenario/Intent} (see Appendix~\ref{app:dataset_characteristics}), exhibit more volatile performance across iterations. In contrast, datasets with more balanced class distributions, (e.g.,~\textsc{Banking77}, \textsc{Crime}, and \textsc{DBpedia}) demonstrate smoother performance curves and generally achieve better results. This disparity can be attributed to the data-driven nature of the process: under-represented classes may not be ranked effectively and, in extreme cases, the ground truth might be excluded from the selected labels altogether. When this occurs, these classes are not predicted by the LLM and cannot be reintroduced in subsequent iterations. The case of \textsc{Amazon Massive Intent} ($k=1.5$) exemplifies this phenomenon, where performance gains are initially rapid but errors gradually accumulate for difficult classes, rather than resolving them.

\textbf{Ranking} As shown in Table~\ref{tab:lsr}, our method also demonstrates improvements even without reduction ($k=Full$) as a substantial portion of the performance gain stems from the label ranking process alone, despite the LLM not being explicitly instructed to consider label order. This effect is most evident with \textsc{Mtop Domain}, where we observe improvements from ranking but none from reduction, given its already small label space ($K=11$). On average, label ranking accounts for 78.6\% of the maximum performance gain, while the reduction of the label space contributes to the remaining 21.4\% (calculated from Table \ref{tab:lsr}, $\frac{\text{Average Gain}_{k=\text{full}}}{\text{Average Gain}_{k=2.0}}$). While ranking proves effective, it differs fundamentally from typical embedding-based ranking approaches~\cite{karpukhin-etal-2020-dense}, which produce static, non-adaptive rankings. In contrast, our classifier is a data-driven ranker which adapts to the task at test-time. These rankings are calibrated across samples and iteratively refined based on the current top prediction from that ranking. For a more detailed analysis and comparison with embedding-based rankings, see Appendix~\ref{app:ranking}.

\textbf{Selection} Our experiments demonstrate that the optimal value of $k$ varies across tasks. We hypothesize that this variation in optimal label space size is related to the semantic similarity between labels. When labels are more semantically similar, a larger label space may be necessary to effectively distinguish between closely related options. This hypothesis is supported by the difference observed in Table~\ref{tab:lsr} between \textsc{Amazon Massive Scenario} and \textsc{Amazon Massive Intent} classifications, where \textsc{Intent}, being a more fine-grained classification of \textsc{Scenario}, requires a larger optimal $k$. Through empirical analysis across datasets, we found that $k$ reaches an optimum at $k=2$ and $k=5$, consistently yielding the highest average performance improvements. The range $k \in [2,5]$ appears to provide a good balance between exploration and exploitation, suggesting these values as practical defaults for zero-shot applications.

\textbf{Majority Voting} The use of majority voting across iterations in our method resembles aspects of Self-Consistency (SC)~\cite{wang2023selfconsistencyimproveschainthought}, which has similarly demonstrated improvements in zero-shot performance at test time. We therefore conducted a comparison with CoT-SC using a temperature of 0.7 and 15 resamples to ensure an equal comparison before majority voting, as shown in Table~\ref{tab:lsr}. While CoT-SC showed a modest average absolute improvement of 1.6\%, our method achieved substantially higher gains of up to 7.0\% on average. Note that these averages are indicative and are highly dependent on the task. We hypothesize that CoT-SC could be complementary to our technique, as its performance gains originate from a different source, i.e., from sampling different reasoning paths of the model with a high temperature~\cite{wang2023selfconsistencyimproveschainthought}. With additional resources, further experiments could evaluate the extent of this complementarity.

\subsubsection{Comparison across Multiple Large Language Models}
We conduct a comprehensive evaluation of LSR with different sized LLMs and capabilities to demonstrate its general applicability. In this setup, we set $k=2$, which we determined to be an optimal value. The results are reported in Table \ref{tab:models} and a figure can also be found in Appendix \ref{app:models}.

\begin{table*}[hbt!]
\caption{Results of LSR with various LLMs. We report the gain in macro-F1 scores between zero-shot predictions and after 15 iterations with $k=2$.}
\label{tab:models}
\begin{center}
\begin{small}
\begin{sc}
\begin{tabular}{l|cccccccc}
\toprule
\begin{tabular}[t]{@{}c@{}} \textbf{Model} \end{tabular} & \begin{tabular}[t]{@{}c@{}} \textbf{Amazon} \\ \textbf{Massive} \\ \textbf{Scenario} \end{tabular} & \begin{tabular}[t]{@{}c@{}} \textbf{Amazon} \\ \textbf{Massive} \\ \textbf{Intent} \end{tabular} & \textbf{Banking77} & \begin{tabular}[t]{@{}c@{}} \textbf{Mtop} \\ \textbf{Domain} \end{tabular} & \begin{tabular}[t]{@{}c@{}} \textbf{Mtop} \\ \textbf{Intent} \end{tabular} & \textbf{Crime} & \textbf{DBpedia} \\
\midrule
Llama-3.1-70B & .719 & .778 & .670 & .945 & .633 & .765 & .555 \\
\quad w/ LSR & +.050 & +.027 & +.129 & +.014 & +.050 & +.077 & +.142 \\
\midrule
Gemma-2-27b & .682 & .793 & .718 & .945 & .629 & .757 & .572 \\
\quad w/ LSR & +.047 & +.011 & +.063 & +.006 & +.041 & +.051 & +.140 \\
\midrule
Qwen2.5-72B & .720 & .806 & .726 & .961 & .674 & .796 & .639 \\
\quad w/ LSR & +.038 & +.012 & +.072 & +.002 & +.021 & +.065 & +.130 \\
\midrule
\begin{tabular}[t]{@{}l@{}} Claude-3.5-Son. \end{tabular} & .753 & .827 & .789 & .954 & .708 & .843 & .669 \\
\quad w/ LSR & +.035 & +.009 & +.027 & +.009 & +.009 & +.030 & +.111 \\
\bottomrule
\end{tabular}
\end{sc}
\end{small}
\end{center}
\end{table*}

Our experiments empirically validate that LSR consistently improves performance across all LLMs and datasets tested. On average, \textsc{Llama-3.1-70B} gains 7.0\%, \textsc{Gemma-2-27b} gains 5.1\%, \textsc{Qwen2.5-72B} gains 4.9\%, and \textsc{Claude-3.5-Sonnet} gains 3.3\% across all datasets. The magnitude of improvement varies, as the potential for improvement is inversely related to the baseline performance on a given task. This can be attributed to task difficulty: for example, all models score 95\% or higher on \textsc{Mtop Domain} and already saturate the benchmark. Additionally, model capability plays a role, as we generally observe smaller performance improvements from more capable models like \textsc{Qwen2.5-72B} and \textsc{Claude-3.5-Sonnet}. A notable exception is \textsc{DBpedia}, where all models gain at least 11\% using LSR. This leads us to hypothesize that more capable models can inherently handle larger label spaces more effectively. Conversely, less capable models show more substantial performance gains from LSR. This finding has important implications, as our method could enable smaller, more accessible models to achieve performance levels comparable to larger, more resource-intensive models.

\subsection{Direct Inference}\label{sec:direct_infer}
We evaluate the direct inference approach outlined in Section \ref{sec:practical_inference} using pseudo-labels accumulated from 15 iterations of \textsc{Llama-3.1-70B-Instruct} with $k=2.0$, we train a \textsc{CatBoost} classifier with a 20\% stratified hold-out set to prevent overfitting.

This classifier demonstrates performance (Table~\ref{tab:direct_inference} in Appendix~\ref{app:direct_inference}) comparable to the offline methodology ($k=2.0$ in Table~\ref{tab:lsr}), namely within~0.5\% absolute points difference, except for \textsc{Mtop Intent} where the classifier shows 2.2\% lower performance. This could be attributed to epistimic uncertainty or noise in the training data, possibly due to inconsistent labeling by the LLM or inadequate feature representations. Interestingly, when making additional LLM predictions using the classifier's rankings, we observe decreased performance across most metrics (Table~\ref{tab:direct_inference}), except for \textsc{Mtop Intent} which now improves, to within 0.6\% of the offline performance. These results show that the classifier actually exceeds the LLM's zero-shot capabilities, as we effectively distilled knowledge from multiple LLM outputs.

While these results are promising for offline evaluation and benchmarking purposes, our goal is to deploy a frozen classifier for inference. Therefore, we evaluate the classifier's performance on a separate test set of equal size and class distribution to simulate a real-world deployment scenario.\\
Our experiments (Unseen Data in Table~\ref{tab:direct_inference} of Appendix~\ref{app:direct_inference}) reveal that the classifier's performance degrades substantially across most datasets in this scenario, though mostly maintaining positive improvements. For instance, the improvement margin decreases from +8.1\% to +2.3\% for \textsc{Crime}, and in the most severe case, decreases from +14.7\% to -3.4\% for \textsc{DBpedia}. This performance degradation can be attributed to two main factors. First, there is a distribution shift. Changes in zero-shot LLM performance indicate that the test data distribution differs from the training data, affecting the classifier's effectiveness. Second, there are generalization challenges. This is particularly evident in datasets with limited samples per class. For example, in \textsc{Mtop Intent}, 41 classes have fewer than 5 samples, including 15 classes with just one sample, making it difficult for the classifier to generalize effectively.

Interestingly, we find that adding an extra LLM inference step helps mitigate performance degradation as it is less susceptible to distribution shift due to extensive pre-training and zero-shot capabilities. This is evidenced by improvements across all datasets, with gains improving from 2.3\% to 5.8\% for \textsc{Crime} and -3.4\% to +6.5\% for \textsc{DBpedia}. While this approach can be effective during a cold-start, we recommend collecting a minimum number of samples per class (using pseudo-labels) to improve the classifier's generalization to unseen data. Nevertheless, challenges related to generalization and distribution shifts are inherent to data-driven approaches and should be carefully considered during real-world deployment.

\section{Conclusion}\label{sec:conclusion}
In this paper we present a novel method for improving zero-shot classification performance of LLMs through iterative ranking and reduction of the label space. Our results demonstrate that optimizing the label space significantly improves LLM accuracy in classification tasks. Although the process is computationally intensive, we address this limitation through distillation, resulting in a cost-effective classifier that outperforms standard LLM zero-shot classification. An interesting direction for future work is to explore the impact of different classifiers and numerical representation models. Furthermore, the current reliance on empirically determined parameters also suggests opportunities for automated parameter selection. Another promising avenue for future research is extending this approach to multi-label classification, which would further generalize the applicability of our method in real-world scenarios. Ultimately, these findings advance the field of zero-shot learning and provide a promising direction for developing more efficient and effective classification systems.

\section*{Acknowledgements}

The funders of the study had no role in study design, data analysis, data interpretation, or writing of the report. This project has received funding from the European Union’s Active and Assisted Living Programme (AAL) via the Emilio project (Grant Agreement No. HBC.2021.1138) and (partially) funded by the Flemish Government (AI Research Program).

\includegraphics[height=1cm]{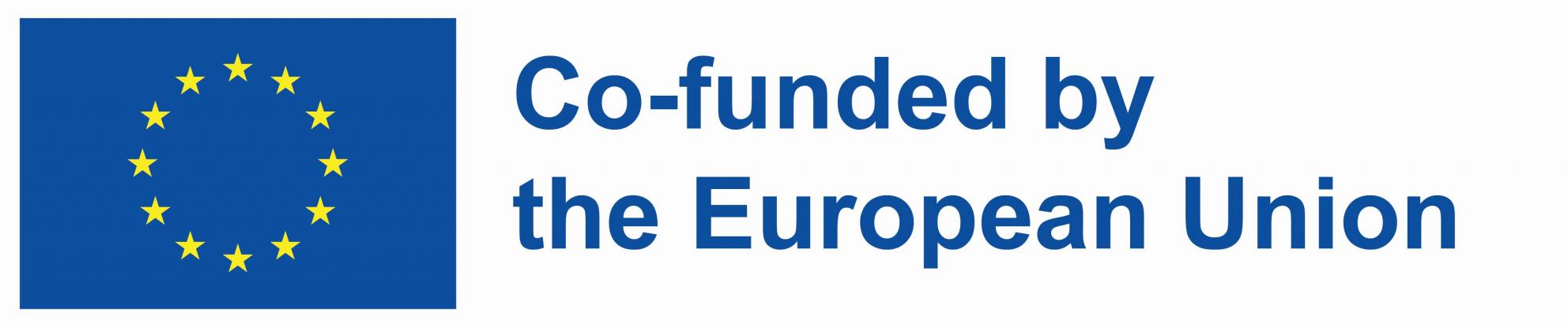} \hspace{0.5em}
\vspace{10pt}

\section*{Impact Statement}
This paper presents work whose goal is to advance the field of 
Machine Learning. There are many potential societal consequences 
of our work, none which we feel must be specifically highlighted here.

\bibliography{main}
\bibliographystyle{icml2025}

\newpage
\appendix
\onecolumn

\section{Dataset Characteristics}\label{app:dataset_characteristics}
We evaluate our method on seven diverse datasets spanning different classification tasks and domains (see Table \ref{tab:dataset_characteristics}). The selection includes standard text classification benchmarks and a tabular classification task, with each dataset containing a substantial number of classes to properly evaluate LSR. Due to budgetary constraints, we create subsets of some of the larger datasets.

\begin{table}[hbt!]
\caption{Characteristics of 7 classification datasets across multiple domains. Key attributes include task type, domain, language, sample size, number of classes, and class balance distribution.}
\label{tab:dataset_characteristics}
\vskip 0.15in
\begin{center}
\begin{small}
\begin{sc}
\begin{tabular}{l|ccccccc}
\toprule
& \begin{tabular}[t]{@{}c@{}} \textbf{Amazon} \\ \textbf{Massive} \\ \textbf{Scenario} \end{tabular} & \begin{tabular}[t]{@{}c@{}} \textbf{Amazon} \\ \textbf{Massive} \\ \textbf{Intent} \end{tabular} & \textbf{Banking77} & \begin{tabular}[t]{@{}c@{}} \textbf{Mtop} \\ \textbf{Domain} \end{tabular} & \begin{tabular}[t]{@{}c@{}} \textbf{Mtop} \\ \textbf{Intent} \end{tabular} & \textbf{Crime} & \textbf{DBpedia} \\
\midrule
Task & \begin{tabular}[c]{@{}c@{}} Text \\ class. \end{tabular} & \begin{tabular}[c]{@{}c@{}} Text \\ class. \end{tabular} & \begin{tabular}[c]{@{}c@{}} Text \\ class. \end{tabular} & \begin{tabular}[c]{@{}c@{}} Text \\ class. \end{tabular} & \begin{tabular}[c]{@{}c@{}} Text \\ class. \end{tabular} & \begin{tabular}[c]{@{}c@{}} Tabular \\ class. \end{tabular} & \begin{tabular}[c]{@{}c@{}} Text \\ class. \end{tabular} \\
Domain & \begin{tabular}[c]{@{}c@{}} Customer \\ Service \end{tabular} & \begin{tabular}[c]{@{}c@{}} Customer \\ Service \end{tabular} & Finance & \begin{tabular}[c]{@{}c@{}} Virtual \\ Assistant \end{tabular} & \begin{tabular}[c]{@{}c@{}} Virtual \\ Assistant \end{tabular} & Legal & \begin{tabular}[c]{@{}c@{}} General \\ Knowledge \end{tabular} \\
Language & English & English & English & English & English & English & English \\
\# Samples & 2974 & 2974 & 3080 & 4386 & 4386 & 3514 & 3448 \\
\# Classes & 18 & 59 & 77 & 11 & 102 & 37 & 70 \\
\begin{tabular}[l]{@{}l@{}} Balanced \\ Classes ? \end{tabular} & No & No & Yes & No & No & Yes & Yes \\
\bottomrule
\end{tabular}
\end{sc}
\end{small}
\end{center}
\vskip -0.1in
\end{table}

\subsection{MTEB Datasets}
Five datasets (\textsc{Amazon Massive Scenario}, \textsc{Amazon Massive Intent}, \textsc{Banking77}, \textsc{Mtop Domain}, \textsc{Mtop Intent}) were selected from the MTEB leaderboard\footnote{\url{https://huggingface.co/spaces/mteb/leaderboard}}, covering text classification in customer service, finance and virtual assistant domains. The MTEB benchmark~\cite{muennighoff2023mtebmassivetextembedding} contains numerous datasets, from which we specifically selected those with a large number of classes (\textgreater10). We report results on the test sets of these datasets. The second test set mentioned in Section~\ref{sec:direct_infer}, is a subset of the MTEB training data.

\subsection{Crime Dataset}
The \textsc{Crime} tabular dataset is derived from the San Francisco Police Department's incident reports\footnote{Police Department Incident Reports: Historical 2003 to May 2018, \url{https://datasf.org/opendata/}}. It consists of a subset of up to 100 randomly selected samples per class (where available) across 37 distinct crime categories and includes seven key features: Category, Description, Day of Week, Police District, Resolution, Address, and Time. This tabular dataset presents a unique challenge in combining textual and categorical information for classification.

\subsection{DBpedia Dataset}
\textsc{DBpedia}~\cite{lehmann2015dbpedia}, derived from structured content extracted from Wikipedia, provides a classification challenge for encyclopedia articles. While a simpler 14-class variant is commonly used in NLP benchmarks~\cite{zhang2015character}, we specifically use a more challenging hierarchical version available on Kaggle\footnote{\url{https://www.kaggle.com/datasets/danofer/dbpedia-classes}} that contains three levels with 9, 70, and 219 classes respectively. Our evaluation uses a balanced subset of the data, containing 50 random samples per class where available, focusing on the second level (L2) of the hierarchical structure which consists of 70 classes. Although the data has an inherent hierarchical structure, our method treats all classes independently and does not utilize these hierarchical relationships.
\clearpage
\section{Comparison of Sampling Strategies}\label{app:sampling}
We analyze the thresholds of different sampling strategies for the label selection process.  In this analysis, we use LLM \textsc{(Llama-3.1-70B)} predictions to train the classifier and obtain probability-based rankings. Performance is measured using Hit@k metric using Equation~\ref{eq:hitk}, which is also referred to as the ``hit rate". The hit rate represents the percentage of instances where the ground truth label appears in the reduced label space, which may vary from sample to sample depending on the strategy.

% https://cran.r-project.org/web/packages/recometrics/vignettes/Evaluating_recommender_systems.html
\begin{equation}\label{eq:hitk}
\text{Hit@k} = \max_{i=1..k} 
\begin{cases} 
1, & r_i \in T \\
0, & \text{otherwise}
\end{cases}
\end{equation}

We compare five strategies:
\begin{itemize}
\item \textbf{Top-k}: Selects a fixed number $k$ of highest probable labels
\item \textbf{Top-p} (nucleus sampling): Includes labels until the cumulative probability exceeds threshold $p$
\item \textbf{Min-p}: Selects labels whose probabilities exceed a fraction of the highest probability
\item \textbf{Min-p+}: Our adaptation of Min-p that includes the most recent LLM prediction
\item \textbf{Min-p+ (w)}: Min-p+ with class-weighted label space
\end{itemize}

\begin{figure}[hbt!]
\vskip 0.2in
\begin{center}
\centerline{\includegraphics[width=0.8\textwidth]{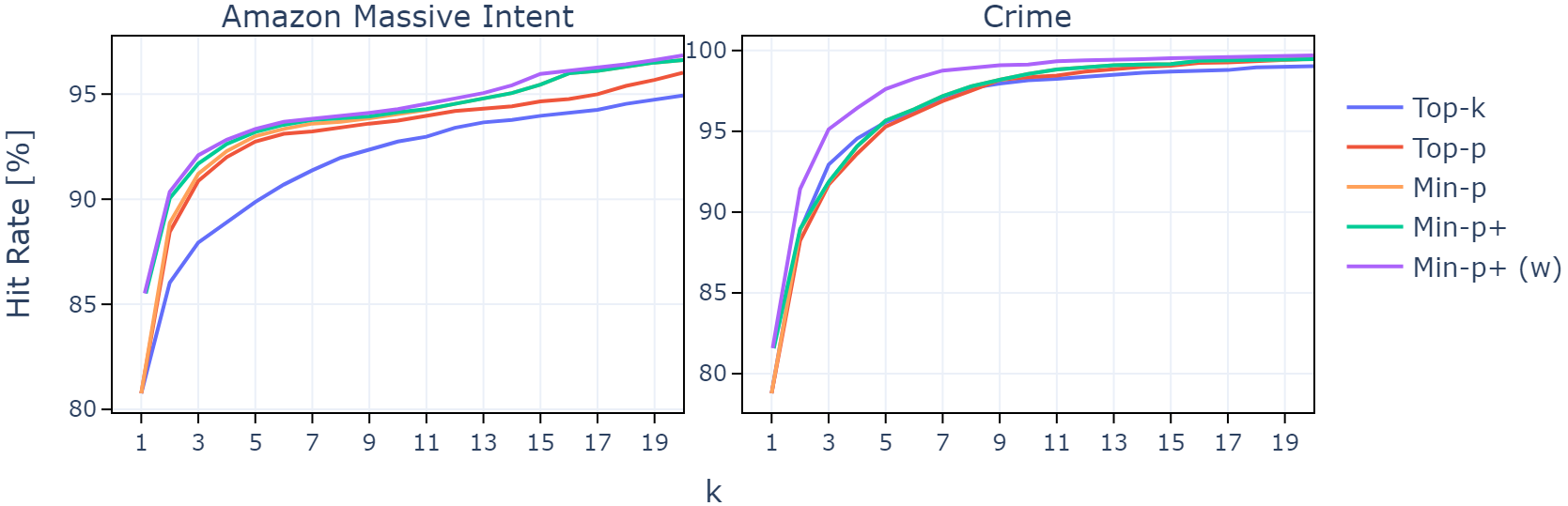}}
\caption{Hit rates of sampling strategies across reduced label space sizes ($k$). Results show the percentage of true labels inside the reduced space using \textsc{Llama-3.1-70B} during the first iteration of LSR.}
\label{fig:sampling_strategy}
\end{center}
\vskip -0.2in
\end{figure}

As illustrated in Figure \ref{fig:sampling_strategy}, common sampling strategies exhibit notable limitations. Top-k sampling, while straightforward, is suboptimal due to its fixed selection size that fails to adapt to varying probability distributions, often including irrelevant low-probability classes. Top-p sampling, though more distribution-aware, is limited by its hard probability cutoff, which can arbitrarily exclude classes with similar probabilities that fall just below the threshold.

Min-p sampling offers a more effective approach to label selection by implementing an adaptive threshold relative to the highest probability. In our methodology, we adapt Min-p sampling with two key modifications. We include the current LLM prediction to increase stability throughout the iteration. Additionally, we introduce class weighting of the label space to address cases where probabilities are more uniformly distributed, as unweighted approaches tend to select many labels in these situations, skewing the average selection size towards challenging cases. By weighting the label space according to class frequencies, we achieve better balance between easy and difficult classes, leading to improved hit rates.
\clearpage
\section{Direct Inference} \label{app:direct_inference}
\begin{table*}[htb!]
\caption{Results of LSR with direct inference. \textsc{Llama-3.1-70B} w/ LSR is distilled into a probabilistic classifier (\textsc{Catboost}). All macro-F1 scores show gains over zero-shot baseline predictions on two different test sets.}
\label{tab:direct_inference}
\begin{center}
\begin{small}
\begin{sc}
\begin{tabular}{l|cccccccc}
\toprule
\begin{tabular}[t]{@{}c@{}} \textbf{Model} \end{tabular} & \begin{tabular}[t]{@{}c@{}} \textbf{Amazon} \\ \textbf{Massive} \\ \textbf{Scenario} \end{tabular} & \begin{tabular}[t]{@{}c@{}} \textbf{Amazon} \\ \textbf{Massive} \\ \textbf{Intent} \end{tabular} & \textbf{Banking77} & \begin{tabular}[t]{@{}c@{}} \textbf{Mtop} \\ \textbf{Domain} \end{tabular} & \begin{tabular}[t]{@{}c@{}} \textbf{Mtop} \\ \textbf{Intent} \end{tabular} & \textbf{Crime} & \textbf{DBpedia} \\
\midrule
Llama-3.1-70B & .719 & .778 & .670 & .945 & .633 & .765 & .555 \\
%\quad w/ LSR & +.050 & +.027 & +.129 & +.014 & +.050 & +.077 & +.142 \\ 
Classifier & \textbf{+.045} & \textbf{+.022} & \textbf{+.136} & \textbf{+.013} & +.028 & \textbf{+.081} & \textbf{+.147} \\
\quad w/ LLM & +.037 & \textbf{+.022} & +.111 & +.004 & \textbf{+.044} & +.072 & +.142
\\
\midrule
\textbf{Unseen Test} & & & & & & & \\
\midrule
Llama-3.1-70B & .721 & .644 & .666 & .945 & .337 & .794 & .570 \\
Classifier (frozen) & +.021 & +.005 & +.109 & -.011 & +.039 & +.023 & -.034 \\
\quad w/ LLM & \textbf{+.047} & \textbf{+.024} & \textbf{+.112} & \textbf{+.003} & \textbf{+.135} & \textbf{+.058} & \textbf{+.065} \\

\bottomrule
\end{tabular}
\end{sc}
\end{small}
\end{center}
\end{table*}
\section{LSR Performance Across Different LLMs} \label{app:models}

\begin{figure}[htb!]
\vskip 0.2in
\begin{center}
\centerline{\includegraphics[width=\columnwidth]{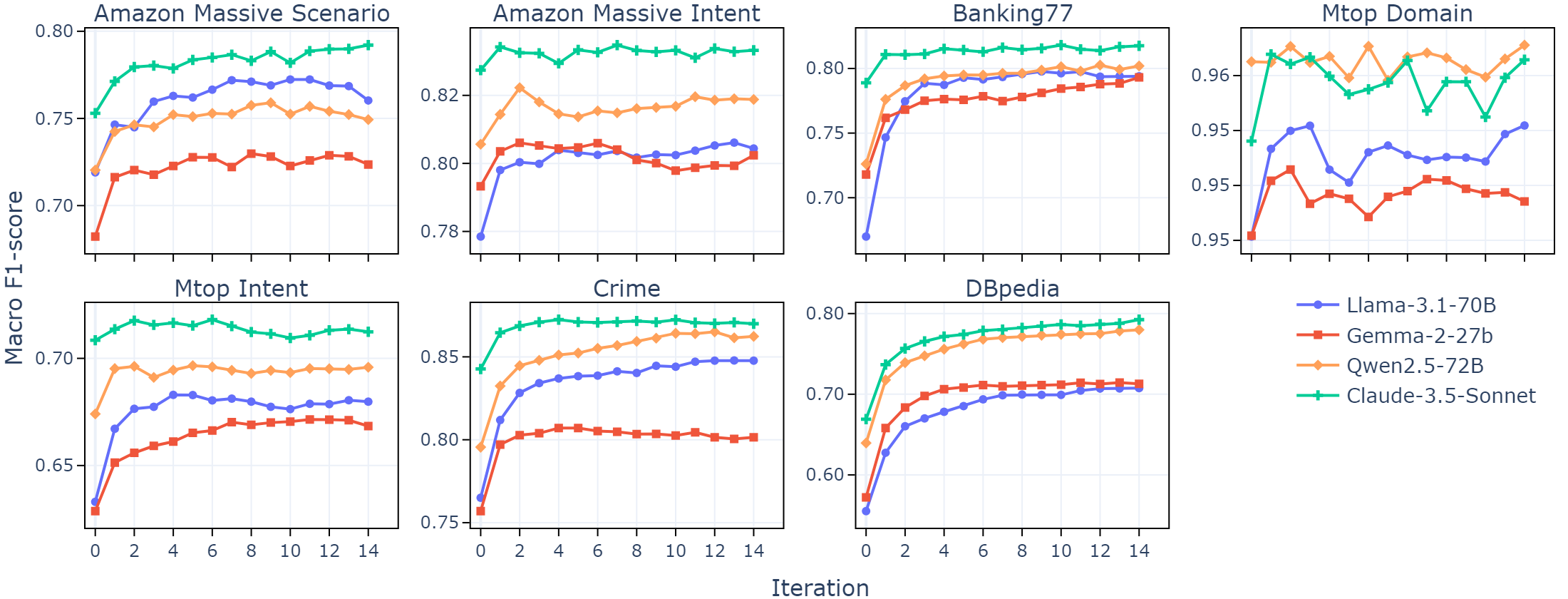}}
\caption{Performance of LSR ($k=2$) with various LLMs. Lines show macro-F1 scores over 15 iterations, starting from zero-shot baseline predictions.}
\label{fig:models}
\end{center}
\vskip -0.2in
\end{figure}

\clearpage
\section{Data Requirement Analysis}
We examine the data requirements by varying the size of the unlabeled dataset used for test-time training, while maintaining the complete test set for consistent evaluation. Using stratified sampling to preserve the original class distributions, we create three representative subsets at 25\%, 50\%, and 75\% of the initial dataset. For each subset, we reserve 20\% as a validation set to prevent overfitting during training, following the methodology.
\begin{figure}[hbt!]
\label{fig:data_size}
\vskip 0.2in
\begin{center}
\centerline{\includegraphics[width=\columnwidth]{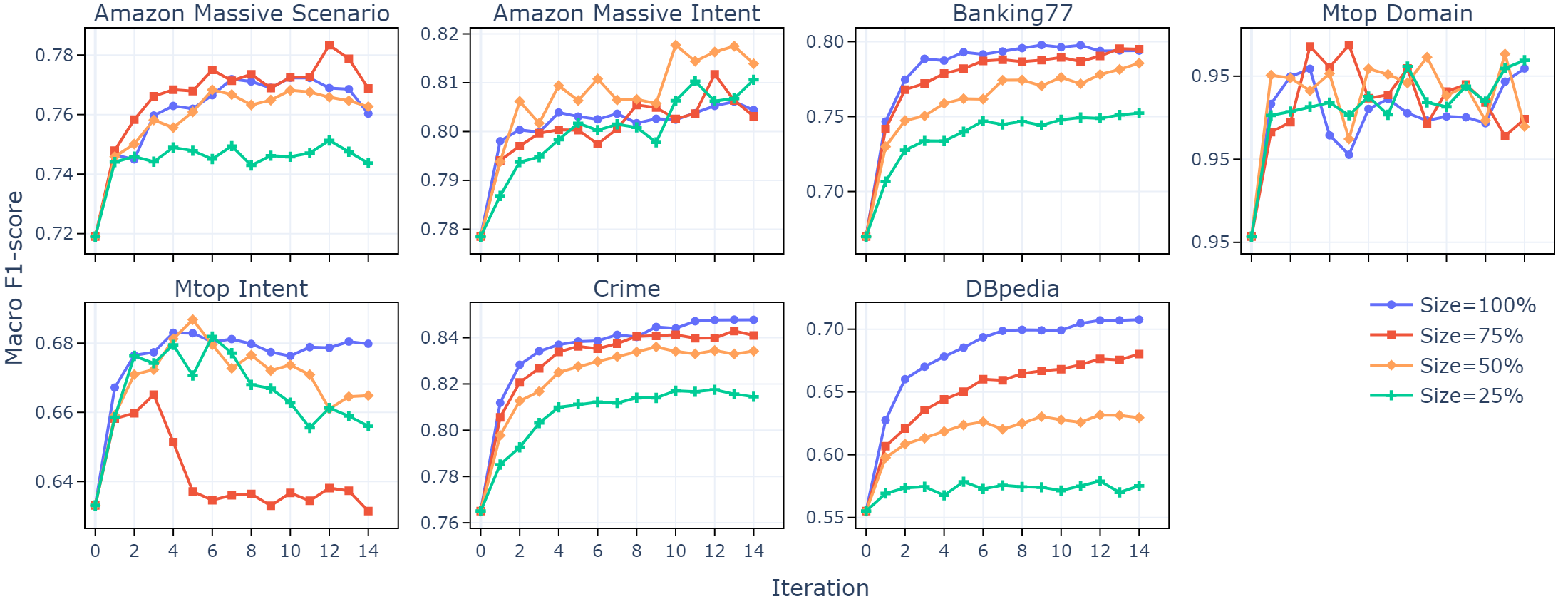}}
\caption{Results of LSR with varying unlabeled subset sizes for training. Lines show macro-F1 scores over 15
iterations, with $k=2$ using \textsc{Llama-3.1-70B}.}
\end{center}
\vskip -0.2in
\end{figure}

\begin{table}[hbt!]
\caption{Effect of data size on LSR. Results show the gain in macro-F1 scores of \textsc{Llama-3.1-70B} between zero-shot predictions and after 15 iterations with $k=2$.}
\label{tab:data_size}
\begin{center}
\begin{small}
\begin{sc}
\begin{tabular}{l|cccccccc}
\toprule
\begin{tabular}[t]{@{}c@{}} \textbf{Data} \\ \textbf{Size} \end{tabular} & \begin{tabular}[t]{@{}c@{}} \textbf{Amazon} \\ \textbf{Massive} \\ \textbf{Scenario} \end{tabular} & \begin{tabular}[t]{@{}c@{}} \textbf{Amazon} \\ \textbf{Massive} \\ \textbf{Intent} \end{tabular} & \textbf{Banking77} & \begin{tabular}[t]{@{}c@{}} \textbf{Mtop} \\ \textbf{Domain} \end{tabular} & \begin{tabular}[t]{@{}c@{}} \textbf{Mtop} \\ \textbf{Intent} \end{tabular} & \textbf{Crime} & \textbf{DBpedia} \\
\midrule
100\% & +.050 & \textbf{+.027} & \textbf{+.129} & +.014 & \textbf{+.050} & \textbf{+.077} & \textbf{+.142} \\
75\% & \textbf{+.060} & \textbf{+.027} & +.122 & +.015 & +.005 & +.076 & +.111 \\
50\% & +.056 & +.041 & +.105 & \textbf{+.017} & +.049 & +.071 & +.072 \\
25\% & +.034 & +.034 & +.082 & \textbf{+.017} & +.046 & +.051 & +.023 \\
\bottomrule
\end{tabular}
\end{sc}
\end{small}
\end{center}
\end{table}

For balanced datasets (\textsc{Banking77}, \textsc{Crime}, \textsc{DBpedia}), we observe a clear positive correlation between data availability and performance improvement, as shown in Table \ref{tab:data_size}. However, the marginal benefits vary across datasets. \textsc{Banking77} and \textsc{Crime} shows diminishing returns when data availability increases from 75\% to 100\%, indicating that our methodology can be effective even with relatively small amounts of data. Notable improvements are observed with as little as 25\% of the full dataset (approximately 10 samples per class). In contrast, \textsc{DBpedia} demonstrates consistent performance gains with increased data availability (see Figure \ref{fig:data_size}), suggesting that larger unlabeled datasets could yield further improvements.

For unbalanced datasets, the relationship between data availability and performance is more complex. When minority classes have very few samples, sometimes just single instances or none after splitting the subset, performance metrics can show significant variability due to poor generalization as highlighted in Section \ref{sec:direct_infer}. While LSR still shows positive improvements in such cases, we recommend ensuring a minimum number of instances per class before applying LSR. The key focus should be on adequately representing the underlying sample distribution rather than achieving a specific total sample size.
\clearpage
\section{Label Space Ranking Comparison}\label{app:ranking}
We compare the zero-shot rankings of the classifier (\textsc{Catboost}) during LSR with $k=Full$ (which ranks without reducing the label space), to that of the embedding model used in the numerical view. For the latter, we obtain rankings by calculating the cosine similarity between the input and label embeddings. We evaluate using the hit rate as defined in Equation~\ref{eq:hitk} and investigate the Top-k candidates.

\begin{figure}[!h]
\label{fig:ranking}
\vskip 0.2in
\begin{center}
\centerline{\includegraphics[width=\columnwidth]{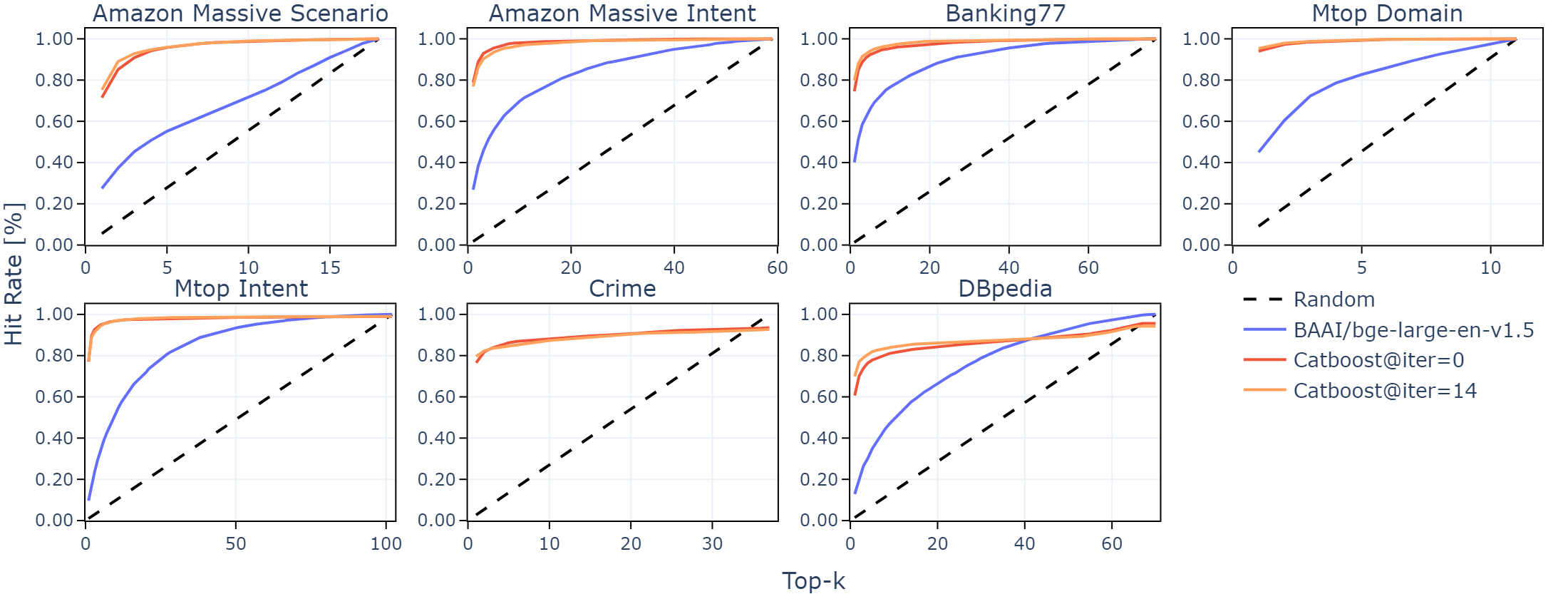}}
\caption{Comparison of zero-shot ranking performance between classifier-based (\textsc{CatBoost}) with LSR and embedding-based (\textsc{BAAI/bge-large-en-v1.5}) approaches across different datasets. The hit rate (Hit@k) is shown for varying numbers of top-k candidates, demonstrating consistently better performance of the classifier-based approach trained on LLM (\textsc{Llama-3.1-70B}) pseudo-labels. Note, the embedding approach is only applicable to text classification tasks.}
\end{center}
\vskip -0.2in
\end{figure}

From Figure~\ref{fig:ranking}, we observe that the classifier-based rankings and classification accuracy (Hit@1) significantly outperform the embedding model across all evaluated datasets.

This finding supports our architectural choice to avoid embedding models for ranking. Since the quality of results would be constrained by the embedding model's initial candidate selection, any errors at this stage would be impossible to recover in subsequent refinement steps. While computationally more intensive, and at risk of retrieval issues associated with long contexts, the ability to process labels within the context window and directly leverage the LLM's knowledge base and reasoning capabilities leads to superior performance.

Figure~\ref{fig:ranking} reveals that the hit rate does not reach 100\% for all datasets when $k$ is equal to the full label space, particularly evident in the DBpedia dataset. This occurs when certain classes are not predicted by the LLM, and thus the classifier cannot incorporate them during training. Furthermore, we observe that while the rankings are properly optimized among the top candidates, this optimization does not extend across all class-labels in the label space, unlike what we observe for the embedding-based rankings. While our current implementation with LSR prioritizes top candidates, future work could focus on improving the consistency of rankings across the entire label space.
\clearpage
\section{Prompt Template}\label{app:prompt_template}

\begin{center}
    Prompt template for general classification tasks
\end{center}
\vspace{-2.5mm}
\noindent\rule{\linewidth}{0.5mm}

\begin{verbatim}
### Context ###
Your goal is to predict the correct category given the context for each case.
The categories are: {labels}

### Instructions ###
1. Write down your thinking in a step-by-step way.
2. You MUST pick one of the suggested categories.
3. Your output must be in JSON format structured as follows: 
   {"predictions": [{"Case": 0, "Analysis": "...", "Label": "..."}, ...]}
4. You must analyze all cases individually.

### Cases ###
Case 0: {feature_value_pairs}, suggestions: {subset_of_labels}
...
(Note: "suggestions" only appear after initial classification)
\end{verbatim}
\rule{\linewidth}{0.5mm}
\textbf{Example input Crime dataset}
\begin{verbatim}
feature_value_pairs:
    Descript: FALSE IMPRISONMENT, DayOfWeek: Sunday, PdDistrict: TARAVAL, 
    Resolution: NONE, Address: 2400 Block of 28TH AV, Hour: 19

subset_of_labels:
    ['KIDNAPPING', 'ASSAULT', 'OTHER OFFENSES', 'SEX OFFENSES, FORCIBLE', 
    'SUSPICIOUS OCC']
\end{verbatim}

\textbf{Example input Banking77 dataset}
\begin{verbatim}
feature_value_pairs:
    text: Can I make sure my card is delivered on a specific day?

subset_of_labels:
    ['card_delivery_estimate', 'card_arrival', 'get_physical_card', 
    'order_physical_card', 'getting_spare_card']
\end{verbatim}

\rule{\linewidth}{0.5mm}

\end{document}